\documentclass[10pt,twocolumn,letterpaper]{article}
\usepackage{booktabs}
\usepackage{multirow}
\usepackage{cvpr}
\usepackage{xcolor}
\usepackage{times}
\usepackage{epsfig}
\usepackage{graphicx}
\usepackage{amsmath}
\usepackage{amssymb}
\usepackage{placeins}
\usepackage{enumitem}

\usepackage[pagebackref=true,breaklinks=true,letterpaper=true,colorlinks,bookmarks=false]{hyperref}

\cvprfinalcopy

\definecolor{tablegreen}{rgb}{0.09, 0.45, 0.27}
\definecolor{tablered}{rgb}{0.55, 0.0, 0.0}
\definecolor{figureviolet}{rgb}{0.74, 0.33, 0.54}

\newcommand{\ith}[1]{{(#1)}}
\newcommand{\iith}{\ith{i}}
\newcommand{\distas}{\sim}
\newcommand{\jpegk}{JPEG\kern0.2ex2000\xspace}

\newcommand{\pmix}{p_m}
\newcommand{\plogistic}{p_\text{L}}

\newcommand{\expectover}[2]{\mathbb{E}_{#1}\left[#2\right]}

\newcommand{\preal}{\tilde p}
\newcommand{\pmodel}{p}
\newcommand{\xl}{x_l}

\makeatletter
\renewcommand{\paragraph}{%
  \@startsection{paragraph}{4}%
  {\z@}{2ex \@plus 1.25ex \@minus .5ex}{-1em}%
  {\normalfont\normalsize\bfseries}%
}
\makeatother

\begin{document}

\title{Learning Better Lossless Compression Using Lossy Compression}

\author{Fabian Mentzer \\
ETH Zurich \\[-0.8ex]
{\scriptsize mentzerf@vision.ee.ethz.ch} 
\and
Luc Van Gool \\
ETH Zurich \\[-0.8ex]
{\scriptsize vangool@vision.ee.ethz.ch} 
\and
Michael Tschannen \\
Google Research, Brain Team \\[-0.8ex]
{\scriptsize tschannen@google.com} 
}

\maketitle

\begin{abstract}
We leverage the powerful lossy image compression algorithm BPG to build a lossless image compression system. Specifically, the original image is first decomposed into the lossy reconstruction obtained after compressing it with BPG and the corresponding residual.
We then model the distribution of the residual with a convolutional neural network-based probabilistic model that is conditioned on the BPG reconstruction, and combine it with entropy coding to losslessly encode the residual. Finally, the image is stored using the concatenation of the bitstreams produced by BPG and the learned residual coder.
The resulting compression system achieves state-of-the-art performance in learned lossless full-resolution image compression, outperforming previous learned approaches as well as PNG, WebP, and JPEG2000.
\end{abstract}

\section{Introduction} \label{sec:intro}

The need to efficiently store the ever growing amounts of data generated continuously on mobile devices has spurred a lot of research on compression algorithms. Algorithms like JPEG~\cite{jpeg1992wallace} for images and H.264~\cite{wiegand2003overview} for videos are used by billions of people daily.

After the breakthrough results achieved with deep neural networks in image classification~\cite{krizhevsky2012imagenet}, and the subsequent rise of deep-learning based methods, \emph{learned} lossy image compression has emerged as an active area of research (e.g.~\cite{balle2016end, theis2017lossy, toderici2016full, rippel17a, agustsson2017soft,agustsson2018generative,minnen2018joint,mentzer2018cvpr,tschannen2018lossy}).
In \emph{lossy} compression, the goal is to achieve small bitrates $R$ given a certain allowed distortion $D$ in the reconstruction, i.e., 
the rate-distortion trade-off $R + \lambda D$ is optimized.
In contrast, in \emph{lossless} compression, no distortion is allowed, and we aim to reconstruct the input perfectly by transmitting as few bits as possible.
To this end, a probabilistic model of the data can be used together with entropy coding techniques to encode and transmit data via a bitstream. The theoretical foundation for this idea is given in Shannon's landmark paper~\cite{shannon1948it}, 
which proves a lower bound for the bitrate achievable by such a probabilistic model, and the overhead incurred by using an imprecise model of the data distribution. One beautiful result is 
that maximizing the likelihood of a parametric probabilistic model is equivalent to minimizing the bitrate obtained when using that model for lossless compression with an entropy coder (see, e.g., \cite{l3c}).
Learning parametric probabilistic models by likelihood maximization has been studied to a great extent in the generative modeling literature (e.g.~\cite{van2016pixel, van2016conditional, Salimans2017pcnnpp, reed2017parallel, kolesnikov2017pixelcnn}). Recent works have linked these results to learned lossless compression~\cite{l3c,hoogeboom2019integer,townsend2019practical,kingma2019bit}.

\begin{figure}[t!]
\centering
\includegraphics[width=\linewidth]{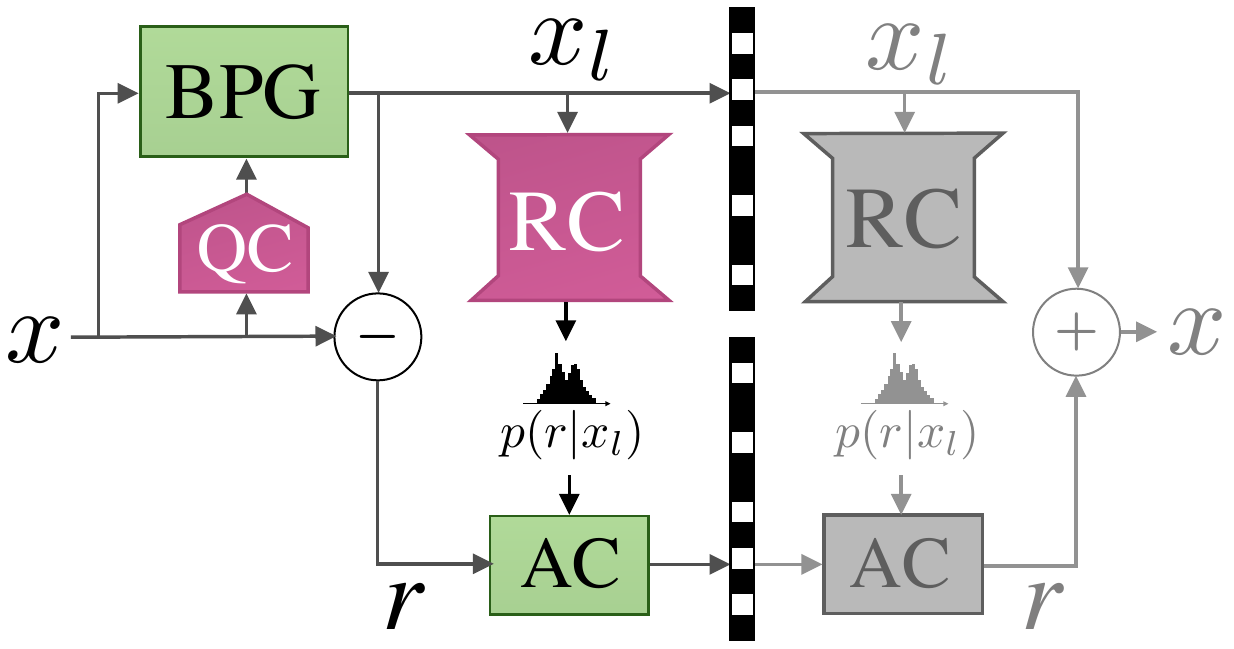}
    \caption{\label{fig:model}Overview of the proposed learned lossless compression approach. To encode an input image $x$, we feed it into the Q-Classifier (QC) CNN to obtain an appropriate quantization parameter $Q$, which is used to compress $x$ with BPG. The resulting lossy reconstruction $x_l$ is fed into the Residual Compressor (RC) CNN, which predicts the probability distribution of the residual, $p(r|x_l)$, conditionally on $x_l$. An arithmetic coder (AC) encodes the residual $r$ to a bitstream, given $p(r|x_l)$. In gray we visualize how to reconstruct $x$ from the bistream. Learned components are shown in \emph{\textcolor{figureviolet}{violet}}.}\vspace{-1ex}
\end{figure}

Even though recent learned lossy image compression methods achieve state-of-the-art results on various data sets, the results obtained by the non-learned H.265-based BPG~\cite{sullivan2012overview,bpg} are still highly competitive,
without requiring sophisticated hardware accelerators such as GPUs to run.
While BPG was outperformed by learning-based approaches across the bitrate spectrum in terms of PSNR~\cite{minnen2018joint} and visual quality~\cite{agustsson2018generative}, it still excels particularly at high-PSNR lossy reconstructions.

In this paper, we propose a learned \emph{lossless} compression system by leveraging the power of the lossy BPG, as illustrated in Fig.~\ref{fig:model}. 
Specifically, we decompose the input image $x$ into the lossy reconstruction $\xl$ produced by BPG and the corresponding residual $r$. We then learn a probabilistic model $p(r|\xl)$ of the residual, conditionally on the lossy reconstruction $\xl$. This probabilistic model is fully convolutional and can be evaluated using a single forward pass, both for encoding and decoding. We combine it with an arithmetic coder to losslessly compress the residual and store or transmit the image as the concatenation of the bitstrings produced by BPG and the residual compressor. Further, we use a computationally inexpensive technique from the generative modeling literature, tuning the ``certainty'' (temperature) of $p(r|\xl)$, as well as an auxiliary shallow classifier to predict the quantization parameter of BPG in order to optimize our compressor on a per-image basis. These components together lead to a state-of-the-art full-resolution learned lossless compression system.
All of our code and data sets are available on github.\footnote{\url{https://github.com/fab-jul/RC-PyTorch}}

In contrast to recent work in lossless compression, we do not need to compute and store any side information (as opposed to L3C~\cite{l3c}), and our CNN is lightweight enough to train and evaluate on high-resolution natural images (as opposed to~\cite{hoogeboom2019integer, kingma2019bit}, which have not been scaled to full-resolution images to our knowledge).

In summary, our main contributions are:
\begin{itemize}[leftmargin=*,parsep=0pt,itemsep=0pt,topsep=1pt]
    \item We leverage the power of the classical state-of-the-art lossy compression algorithm BPG in a novel way to build a conceptually simple learned lossless image compression system.
    \item Our system is optimized on a per-image basis with a light-weight post-training step, where we obtain a lower-bitrate probability distribution by adjusting the confidence of the predictions of our probabilistic model.
    \item Our system outperform the state-of-the-art in learned lossless full-resolution image compression, L3C \cite{l3c}, as well as the classical engineered algorithms WebP, JPEG200, PNG. Further, in contrast to L3C, we are also outperforming FLIF on Open Images, the domain where our approach (as well as L3C) is trained.
\end{itemize}

\section{Related Work}

\paragraph{Learned Lossless Compression}
Arguably most closely related to this paper, Mentzer~\etal~\cite{l3c} build a computationally cheap hierarchical generative model (termed L3C) to enable practical compression on full-resolution images. 

Townsend~\etal~\cite{townsend2019practical} and Kingma~\etal~\cite{kingma2019bit} leverage the ``bits-back scheme''~\cite{hinton1993keeping} for lossless compression of an image stream, where the overall bitrate of the stream is reduced by leveraging previously transmitted information.
Motivated by recent progress in generative modeling using (continuous) flow-based models (e.g.~\cite{rezende2015variational,kingma2016improved}), Hoogeboom~\etal~\cite{hoogeboom2019integer} propose Integer Discrete Flows (IDFs), defining an invertible transformation for discrete data. 
In contrast to L3C, the latter works focus on smaller data sets such as
MNIST, CIFAR-10, ImageNet32, and ImageNet64, where they 
achieve state-of-the-art results.

\paragraph{Likelihood-Based Generative Modeling}
As mentioned in Section~\ref{sec:intro}, virtually every generative model can be used for lossless compression, when used with an entropy coding algorithm. Therefore, while the following generative approaches do not take a compression perspective, they are still related.
The state-of-the-art PixelCNN~\cite{van2016pixel}-based models rely on auto-regression in RGB space to efficiently model a conditional distribution. The original PixelCNN~\cite{van2016pixel} and PixelRNN~\cite{van2016conditional} model the probability distribution of a pixel given all previous pixels (in raster-scan order). To use these models for lossless compression, $O(H\cdot W)$ forward passes are required, where $H$ and $W$ are the image height and width, respectively. Various speed optimizations and a probability model amendable to faster training were proposed in~\cite{Salimans2017pcnnpp}. Different other parallelization techniques were developed, including those from~\cite{reed2017parallel}, modeling the image distribution conditionally on subsampled versions of the image, as well as those from~\cite{kolesnikov2017pixelcnn}, conditioning on a RGB pyramid and grayscale images. Similar techniques were also used by~\cite{chen2017pixelsnail, parmar2018image}. 

\paragraph{Engineered Lossless Compression Algorithms}
The wide-spread PNG~\cite{pngurl} applies simple autoregressive filters to remove redundancies from the RGB representation (e.g. replacing pixels with the difference to their left neighbor), and then uses the DEFLATE~\cite{deutsch1996deflate} algorithm for compression. In contrast, WebP~\cite{webpurl} uses larger windows to transform the image (enabling patch-wise conditional compression), and relies on a custom entropy coder for compression. Mainly in use for lossy compression, \jpegk~\cite{skodras2001jpeg2000} also has a lossless mode, where an invertible mapping from RGB to compression space is used.
At the heart of FLIF~\cite{flif2016} is an entropy coding method called ``meta-adaptive near-zero integer arithmetic coding'' (MANIAC), which is based on the CABAC method used in, e.g., H.264~\cite{wiegand2003overview}. In CABAC, the context model used to compress a symbol is selected from a finite set based on local context~\cite{richardson2004h}. The ``meta-adaptive'' part in MANIAC refers to the context model which is a decision tree learned \emph{per image}.

\paragraph{Artifact Removal}
Artifact removal methods in the context of lossy compression are related to our approach in that they aim to make predictions about the information lost during the lossy compression process. In this context, the goal is to produce sharper and/or more visually pleasing images given a lossy reconstruction from, e.g., JPEG. Dong~\etal~\cite{dong2015compression} proposed the first CNN-based approach using a network inspired by super-resolution networks. 
\cite{svoboda2016compression} extends this using a residual structure, and~\cite{cavigelli2017cas} relies on hierarchical skip connections and a multi-scale loss. 
Generative models in the context of artifact removal are explored by \cite{galteri2017deep}, which proposes to use GANs~\cite{goodfellow2014generative} to obtain more visually pleasing results. 

\section{Background}

\subsection{Lossless Compression}
\label{sec:losslesscomp}

We give a very brief overview of lossless compression basics here and refer to the information theory literature for details~\cite{shannon1948it,cover2012elements}. In lossless compression, we consider a stream of symbols $x_1, \dots, x_N$, where each $x_i$ is an element from the same finite set $\mathcal{X}$. The stream is obtained by drawing each symbol $x_i$ independently from the same distribution $\preal$, i.e., the $x_i$ are i.i.d.\ according to $\preal$. 
We are interested in encoding the symbol stream into a bitstream, such that we can recover the exact symbols by decoding. In this setup, the \emph{entropy} of $\preal$ is equal to the expected number of bits needed to encode each $x_i$:
\[
    H(\preal) = \text{bits}(x_i) = \expectover{x_i \distas \preal}{-\log_2 \preal(x_i)}.
\]
In general, however, the exact $\preal$ is unknown, and we instead consider the setup where we have an approximate model $\pmodel$. 
Then, the expected bitrate will be equal to the \emph{cross-entropy} between $\preal$ and $\pmodel$, given by:
\begin{equation}
    H(\preal, \pmodel) = \expectover{x_i \distas \preal}{-\log_2 \pmodel(x_i)}. \label{eq:crossentropy}
\end{equation}
Intuitively, the higher the discrepancy between the model $\pmodel$ used for coding is from the real $\preal$, the more bits we need to encode data that is actually distributed according to $\preal$.

\paragraph{Entropy Coding} Given a symbol stream $x_i$ as above and a probability distribution $p$ (not necessarily $\preal$), we can encode the stream using \emph{entropy coding}. Intuitively, we would like to build a table that maps every element in $\mathcal{X}$ to a bit sequence, such that $x_i$ gets a short sequence if $p(x_i)$ is high. The optimum is to output
$\log_2 p(x_i)$ bits for symbol $x_i$, which is what entropy coding algorithms achieve.
Examples include Huffman coding~\cite{huffman1952method} and arithmetic coding~\cite{witten1987arithmetic}. 

In general, we can use a different distribution $p_i$ for every symbol in the stream, as long as the $p_i$ are also available for decoding. \emph{Adaptive} entropy coding algorithms work by allowing such varying distributions as a function of previously encoded symbols. In this paper, we use adaptive arithmetic coding~\cite{witten1987arithmetic}.

\subsection{Lossless Image Compression with CNNs}
\label{sec:losslesscompcnn}

As explained in the previous section, all we need for lossless compression is a model $p$, 
since we can use entropy coding to encode and decode any input losslessly given $p$.
In particular, we can use a CNN to parametrize $p$. To this end, one general approach is
to introduce (structured) side information $z$ available both at encoding and decoding time, and 
model 
the probability distribution of natural images $x^\iith$ conditionally on $z$, using the CNN to parametrize $p(x|z)$.%
\footnote{We write $p(x)$ to denote the entire probability mass function 
and $p(x^\iith)$ to denote $p(x)$ evaluated at $x^\iith$.}
Assuming that both the encoder and decoder have access to $z$ and $p$, we can losslessly encode $x^\iith$ as follows:
We first use the CNN to produce $p(x|z)$. Then, we employ an entropy encoder (described in the previous section) with $p(x|z)$ to encode
$x^\iith$ to a bitstream. To decode, we once again feed $z$ to the CNN, obtaining $p(x|z)$, and decode $x^\iith$ from the bitstream using the entropy decoder.

One key difference among the approaches in the literature is the factorization of $p(x|z)$. In the original PixelCNN paper~\cite{van2016conditional} the image $x$ is modeled as a sequence of pixels, and $z$ corresponds to all previous pixels. Encoding as well as decoding are done autoregressively. In IDF~\cite{hoogeboom2019integer}, $x$ is mapped to a $z$ using an invertible function, and $z$ is then encoded using a fixed prior $p(z)$, i.e., $p(x|z)$ here is a deterministic function of $z$. 
In approaches based on the bits-back paradigm~\cite{townsend2019practical,kingma2019bit}, while encoding, $z$ is obtained by  decoding from additional available information (e.g.\ previously encoded images).
In L3C~\cite{l3c}, $z$ corresponds to features extracted with a hierarchical model that are also saved to the bitstream using hierarchically predicted distributions.

\subsection{BPG}
\label{sec:bpg}

\begin{figure}
\centering
\includegraphics[width=.7\linewidth]{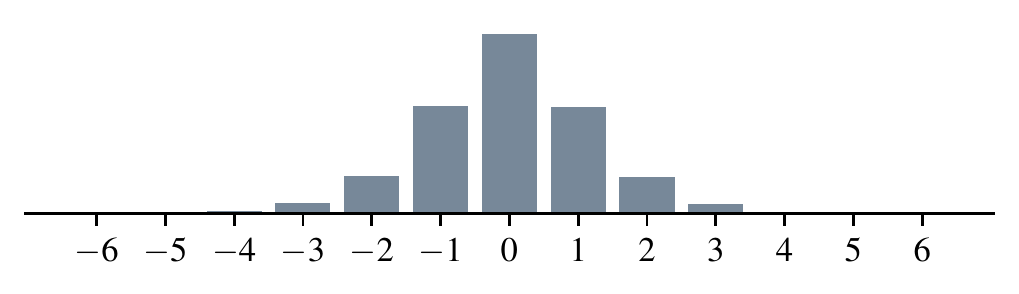}
    \caption{\label{fig:bpg_histo}Histogram of the marginal pixel distribution of residual values obtained using BPG and $Q$ predicted from QC, on Open Images.}
    \vspace{-2ex}
\end{figure}

BPG is a lossy image compression method based on the HEVC video coding standard~\cite{sullivan2012overview}, essentially applying HEVC on a single image. 
To motivate our usage of BPG, we 
show the histogram of the marginal pixel distribution of the residuals obtained by BPG on Open Images (one of our testing sets, see Section~\ref{sec:datasets}) in Fig.~\ref{fig:bpg_histo}. 
Note that while the possible range of a residual is $\{-255, \dots, 255\}$, we observe that for most images, nearly every point in the residual is in the restricted set $\{-6, \dots, 6\}$
, which is indicative of the high-PSNR nature of BPG. 
Additionally, Fig.~\ref{fig:bpg_perf} (in the suppl.) presents a comparison of BPG to the state-of-the-art learned image compression methods, showing that BPG is still very competitive in terms of PSNR.

BPG follows JPEG in having a chroma format parameter to enable color space subsampling, which we disable by setting it to $4{:}4{:}4$.
The only remaining parameter to set is the \emph{quantization parameter} $Q$, where $Q \in \{1, \dots, 51\}$. Smaller $Q$ results in less quantization and thus better quality (i.e., different to the \emph{quality} factor of JPEG, where larger means better reconstruction quality). We learn a classifier to predict $Q$, described in Section~\ref{sec:qclf}.

\begin{figure*}[ht!]
\centering
\includegraphics[width=0.9\textwidth]{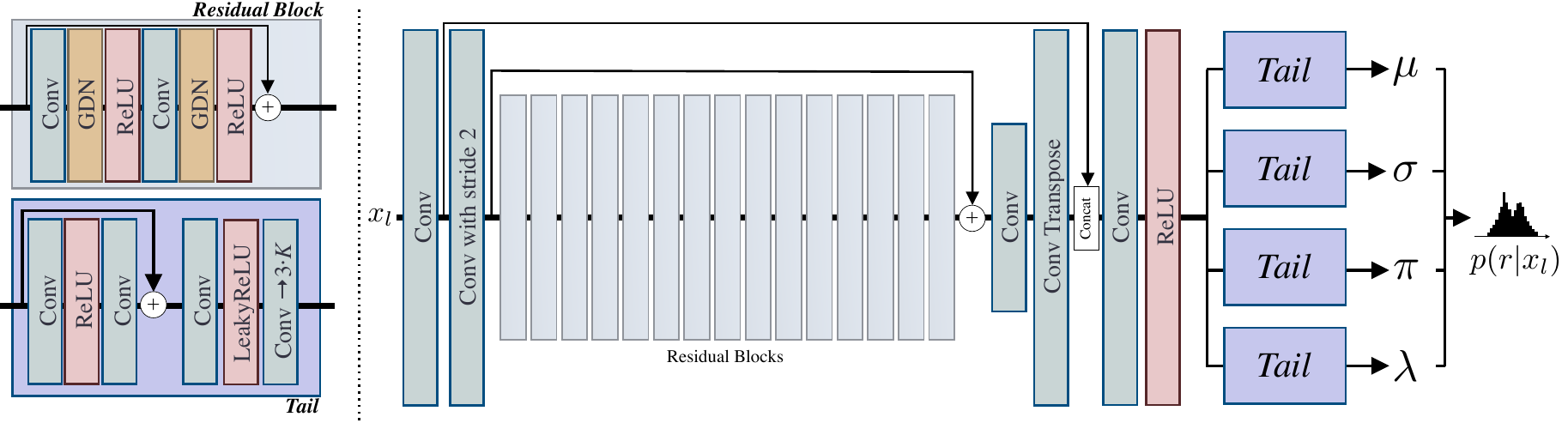}
\caption{\label{fig:archdetail}  The architecture of the residual compressor (RC). On the left, we show a zoom-in of the \emph{Residual Block} and the \emph{Tail} networks. Given $\xl$, the lossy reconstruction of the image $x$, the network predicts the probability distribution of the residual, $p(r|\xl)$. This distribution is a mixture of logistics parametrized via $\mu, \sigma, \pi, \lambda$.
\vspace{-1.3em}
}
\end{figure*}

\section{Proposed Method}

We give an overview of our method in Fig.~\ref{fig:model}.
To encode an image $x$, we first obtain the quantization parameter $Q$ from the \emph{Q-Classifier} (QC) network (Section~\ref{sec:qclf}).
Then, we compress $x$ with BPG, to obtain the lossy reconstruction $\xl$, which we save to a bitstream. Given $\xl$, the \emph{Residual Compressor} (RC) network (Section~\ref{sec:rc}) predicts the probability mass function of the residual $r = x - \xl$, i.e.,
\[
p(r|\xl) = \text{RC}(\xl).
\]
We model $p(r|\xl)$ as a discrete mixture of logistic distributions (Section~\ref{sec:dmol}).
Given $p(r|\xl)$ and $r$, we compress $r$ to the bitstream using adaptive arithmetic coding algorithm (see Section~\ref{sec:losslesscomp}). Thus, the bitstream $B$ consists of the concatenation of the codes corresponding to $x_l$ and $r$.
To decode $x$ from $B$, we first obtain $\xl$ using the BPG decoder, then we obtain once again $p(r|\xl) = \text{RC}(\xl)$, and subsequently decode $r$ from the bitstream using $p(r|\xl)$. Finally, we can reconstruct $x = \xl + r$. In the formalism of Section~\ref{sec:losslesscompcnn}, we have $x=r, z=\xl$. 

Note that no matter how bad RC is at predicting the real distribution of $r$, we can always do \emph{lossless} compression. Even if RC were to predict, e.g., a uniform distribution---in that case, we would just need many bits to store $r$.

\subsection{Residual Compressor} \label{sec:rc}

We use a CNN inspired by ResNet~\cite{he2016deep} and U-Net~\cite{ronneberger2015u}, shown in detail in Fig.~\ref{fig:archdetail}. We first extract an initial feature map $f_\text{in}$ with $C_f=128$ channels, which we then downscale using a stride-2 convolution, and feed through 16 residual blocks. Instead of BatchNorm~\cite{ioffe2015batch} layers as in ResNet, our residual blocks contain GDN layers proposed by~\cite{balle2015density}. Subsequently, we upscale back to the resolution of the input image using a transposed convolution. The resulting features are concatenated with $f_\text{in}$, and convolved to contract the $2\cdot C_f$ channels back to $C_f$, like in U-Net.
Finally, the network splits into four tails, predicting the different parameters of the mixture model, $\pi, \mu, \sigma, \lambda$, described next.

\subsection{Logistic Mixture Model} \label{sec:dmol}
We use a discrete mixture of logistics to model the probability mass function of the residual, $p(r|\xl)$, similar to~\cite{l3c, Salimans2017pcnnpp}. We closely follow the formulation of~\cite{l3c} here:
Let $c$ denote the RGB channel and $u,v$ the spatial location. 
We define 
\begin{equation}
    p(r|x_l) = \prod_{u,v} p(r_{1uv},r_{2uv},r_{3uv}|x_l). \label{p_x_factorization}
\end{equation}
We use a (weak) autoregression over the three RGB channels to define the joint distribution over channels via logistic mixtures $\pmix$:
\begin{align}
    p(r_1, r_2, r_3 | x_l) =\;&\pmix(r_1 | x_l) \cdot \pmix(r_2 | x_l, r_1) \; \cdot \nonumber \\
                                 &\pmix(r_3 | x_l, r_2, r_1),
    \label{eq:p_joint}
\end{align}
where we removed the indices $uv$ to simplify the notation.
    For the mixture $\pmix$ we use a mixture of $K=5$ logistic distributions $\plogistic$. Our distributions are defined by the outputs of the RC network, which yields mixture weights
    $\pi^k_{cuv}$, means $\mu^k_{cuv}$, variances $\sigma^k_{cuv}$, as well as mixture coefficients $\lambda^k_{cuv}$.
    The autoregression over RGB channels is only used to update the means using a linear combination of $\mu$ and the target $r$ of previous channels, scaled by the coefficients $\lambda$. We thereby obtain $\tilde \mu$:
\begin{align}
    \tilde \mu^k_{1uv} = \mu^k_{1uv} \nonumber \hspace{5em}
    &\tilde \mu^k_{2uv} = \mu^k_{2uv} + \lambda^k_{\alpha uv} \; r_{1uv} \nonumber  \\
    \tilde \mu^k_{3uv} = \mu^k_{3uv} + \lambda^k_{\beta uv} &\; r_{1uv} + \lambda^k_{\gamma uv} \; r_{2uv}.
    \label{eq:updating_mus}
\end{align}

With these parameters, we can define
\begin{align}
    \pmix(r_{cuv} | x_l, r_{\text{prev}}) &= \sum_{k=1}^K \pi^k_{cuv} \; \plogistic(r_{cuv} | \tilde \mu^k_{cuv}, \sigma^k_{cuv}), 
    \label{eq:p_mix}
\end{align}
where $r_\text{prev}$ denotes the channels with index smaller than $c$ (see Eq.~\ref{eq:p_joint}), used to obtain $\tilde \mu$ as shown above, and $\plogistic$ is the logistic distribution:
\[
\plogistic(r|\mu,\sigma)=\frac{e^{-(r-\mu)/\sigma}}{\sigma ( 1 + e^{-(r-\mu)/\sigma})^2}.
\]
We evaluate $p_\text{L}$ at discrete $r$, via its CDF, as in~\cite{Salimans2017pcnnpp,l3c},
evaluating
\begin{align}
    \plogistic(r) = \text{CDF}(r + 1/2) -
        \text{CDF}(r - 1/2). \label{eq:plog}
\end{align}

\subsection{Loss}

As motivated in Section~\ref{sec:losslesscomp}, we are interested in minimizing the cross-entropy between the real distribution of the residual 
$\tilde p(r)$ 
and our model $p(r)$: the smaller  the cross-entropy, the closer $p$ is to $\preal$, and the fewer bits an entropy coder will use to encode $r$.
We consider the setting where we have $N$ training images $x^{(1)}, \dots, x^{(N)}$.
For every image, we compute the lossy reconstruction $x_l^{(i)}$ as well as the corresponding residual $r^{(i)} = x^{(i)} - x_l^{(i)}$.
While the true distribution $\tilde p(r)$ is unknown,
we can consider the empirical distribution obtained from the samples and minimize:
\begin{align}
\mathcal L(RC) =-\sum_{i=1}^N \log p(r^{(i)}|x_l^{(i)}). \label{eq:loss}
\end{align}
This loss decomposes over samples, allowing us to minimize it over mini-batches.
Note that minimizing Eq.~\ref{eq:loss} is the same as maximizing the likelihood of $p$, which is the perspective taken in the likelihood-based generative modeling literature.

\subsection{Q-Classifier} \label{sec:qclf}

A random set of natural images is expected to contain images of varying ``complexity'', where complex can mean a lot of high frequency structure and/or noise. 
While virtually all lossy compression methods have a parameter like BPG's $Q$, to navigate the trade-off between bitrate and quality, it is important to note that compressing a random set of natural images with the same fixed $Q$ will usually lead to the bitrates of these images being spread around some $Q$-dependent mean. Thus, in our approach, it is suboptimal to fix $Q$ for all images.

Indeed, in our pipeline we have a trade-off between the bits allocated to BPG and the bits allocated to encoding the residual. This trade-off can be controlled with $Q$: For example, if an image contains components that are easier for the RC network to model, it is beneficial to use a higher $Q$, such that BPG does not waste bits encoding these components. We observe that for a fixed image, and a trained RC, there is a single optimal $Q$.

To efficiently obtain a good $Q$, we train a simple classifier network, the Q-Classifier (QC), and then use
$Q = \text{QC}(x)$
to compress $x$ with BPG.
For the architecture, we use a light-weight ResNet-inspired network with 8 residual blocks for QC, and train it to predict a class in $\mathcal{Q} = \{11, \dots, 17\}$, given an image $x$ ($\mathcal{Q}$ was selected using the Open Images validation set).  In contrast to ResNet, we employ no normalization layers (to ensure that the prediction is independent of the input size). Further, the final features are obtained by average pooling each of the final $C_f=256$ channels of the $C_f{\times}H'{\times}W'$-dimensional feature map. The resulting $C_f$-dimensional vector is fed to a fully connected layer, to obtain the logits for the $|\mathcal{Q}|$ classes, which are then normalized with a softmax.
Details are shown in Section~\ref{sec:qcarch} in the supplementary material.

While the input to QC is the full-resolution image, the network is shallow and downsamples multiple times, making this a computationally lightweight component. 

\subsection{$\tau$-Optimization}\label{sec:tauotpim}

Inspired by the temperature scaling employed in the generative modeling literature (e.g.~\cite{kingma2018glow})
, we further optimize the predicted distribution $p$ with a simple trick: Intuitively, if RC predicts a $\mu_{cuv}$ that is close to the target $r_{cuv}$, we can make the cross-entropy in Eq.~\ref{eq:loss} (and thus the bitrate) smaller by making the predicted logistic ``more certain'' by choosing a smaller $\sigma$. This shifts probability mass towards $r_{cuv}$. However, there is a breaking point, where we make it ``too certain'' (i.e., the probability mass concentrates too tightly around $\mu_{cuv}$) and the cross-entropy increases again.

While RC is already trained to learn a good $\sigma$, the prediction is only based on $x_l$. We can improve the final bitrate during encoding, when we additionally have access to the target $r_{cuv}$, by \emph{rescaling} the predicted $\sigma^k_{cuv}$ with a factor $\tau^k_c$, chosen for every mixture $k$ and every channel $c$. This yields a more optimal
$\tilde \sigma^k_{cuv} = \tau_{c}^k \cdot \sigma^k_{cuv}. $
Obviously, $\tau$ also needs to be known for decoding, and we thus have to transmit it via the bitstream. However, since we only learn a $\tau$ for every channel and every mixture (and not for every spatial location), this causes a completely negligible overhead of $C \cdot K = 3 \cdot 5$ floats $=60$ bytes.

We find $\tau_c^k$ for a given image $x^{(i)}$ by minimizing the likelihood in Eq.~\ref{eq:loss} \emph{on that image}, i.e., we optimize
\begin{align}
\min_\tau \sum_{c,u,v} \log p_\tau(r^{(i)}|x_{l,cuv}^{(i)}), \label{eq:loss_tau}
\end{align}
where $p_\tau$ is equal to $p$ predicted from RC but using $\tilde \sigma_{cuv}^k$. To optimize Eq.~\ref{eq:loss_tau}, we 
use stochastic gradient descent with a very high learning rate of $9\textsc{e}^{-2}$ and momentum $0.9$, which converges in 10-20 iterations, depending on the image.

We note that this is also computationally cheap.
Firstly, we only need to do the forward pass through RC once, to get $\mu, \sigma, \lambda, \pi$, and then in every step of the $\tau$-optimization, we only need to evaluate $\tau_c^k \cdot \sigma_{cuv}^k$ and subsequently Eq.~\ref{eq:loss_tau}. Secondly, the optimization is only over 15 parameters. Finally, since for practical $H{\times}W$-dimensional images, $15 \ll H\cdot W$, we can do the sum in Eq.~\ref{eq:loss_tau} over a $4{\times}$ spatially subsampled version of $p_\tau$.

\begin{table*}[ht!]
\centering
    \begin{tabular}{l@{\hskip 6ex}r@{\hskip 1mm}l@{\hskip 8ex}r@{\hskip 1mm}l@{\hskip 8ex}r@{\hskip 1mm}l@{\hskip 8ex}r@{\hskip 1mm}l}
\toprule
\footnotesize{[bpsp]} & \multicolumn{2}{@{}l}{Open Images} & \multicolumn{2}{@{}l}{CLIC.mobile} & \multicolumn{2}{@{}l}{CLIC.pro} & \multicolumn{2}{@{}l}{DIV2K}
 \\ \midrule

RC (Ours) & 2.790 & & 2.538 & & 2.933 & & 3.079 &\\
L3C & 2.991 & \textcolor{tablegreen}{\footnotesize $+7.2\%$} & 2.639 & \textcolor{tablegreen}{\footnotesize $+4.0\%$} & 2.944 & \textcolor{tablegreen}{\footnotesize $+0.4\%$} & 3.094 & \textcolor{tablegreen}{\footnotesize $+0.5\%$}\\
PNG & 4.005 & \textcolor{tablegreen}{\footnotesize $+44\%$} & 3.896 & \textcolor{tablegreen}{\footnotesize $+54\%$} & 3.997 & \textcolor{tablegreen}{\footnotesize $+36\%$} & 4.235 & \textcolor{tablegreen}{\footnotesize $+38\%$}\\
\jpegk & 3.055 & \textcolor{tablegreen}{\footnotesize $+9.5\%$} & 2.721 & \textcolor{tablegreen}{\footnotesize $+7.2\%$} & 3.000 & \textcolor{tablegreen}{\footnotesize $+2.3\%$} & 3.127 & \textcolor{tablegreen}{\footnotesize $+1.6\%$}\\
WebP & 3.047 & \textcolor{tablegreen}{\footnotesize $+9.2\%$} & 2.774 & \textcolor{tablegreen}{\footnotesize $+9.3\%$} & 3.006 & \textcolor{tablegreen}{\footnotesize $+2.5\%$} & 3.176 & \textcolor{tablegreen}{\footnotesize $+3.2\%$}\\
FLIF & 2.867 & \textcolor{tablegreen}{\footnotesize $+2.8\%$} & 2.492 & \textcolor{tablered}{\footnotesize $-1.8\%$} & 2.784 & \textcolor{tablered}{\footnotesize $-5.1\%$} & 2.911 & \textcolor{tablered}{\footnotesize $-5.5\%$}\\ \bottomrule
\vspace{-1em}
\end{tabular}
\caption{\label{table:results_bpsp}
Compression performance of the proposed method (RC) compared to the learned L3C \cite{l3c}, as well as the classical engineered approaches PNG, \jpegk, WebP, and FLIF.
We show the difference in percentage to our approach, using \emph{\textcolor{tablegreen}{green}}  to indicate that we achieve a better bpsp and  
\emph{\textcolor{tablered}{red}} otherwise.
}
\end{table*}

\section{Experiments}

\subsection{Data sets} \label{sec:datasets}

\paragraph{Training}
Like L3C~\cite{l3c}, we train on $300\,000$ images from the Open Images data set~\cite{OpenImages}. These images are made available as JPEGs, which is not ideal for the lossless compression task we are considering, but we are not aware of a similarly large scale lossless training data set. To prevent overfitting on JPEG artifacts, we downscale each training image using a factor randomly selected from $[0.6, 0.8]$ by means of the Lanczos filter provided by the Pillow library~\cite{pillowurl}.
For a fair comparison, the L3C baseline results were also obtained by training on the exact same data set.

\paragraph{Evaluation}
We evaluate our model on four data sets: \textbf{Open Images} is a subset of 500 images from Open Images validation set, preprocessed like the training data. \textbf{CLIC.mobile} and \textbf{CLIC.pro} are two new data sets
commonly used in recent image compression papers,
released as part of the ``Workshop and Challenge on Learned Image Compression'' (CLIC)~\cite{clicurl}.
CLIC.mobile contains 61 images taken using cell phones, while CLIC.pro contains 41 images from DSLRs, retouched by professionals.
Finally, we evaluate on the 100 images from \textbf{DIV2K}~\cite{agustssondiv2k}, a super-resolution data set with high-quality images. We show examples from these data sets in Section~\ref{sec:testsetexamples}.

For a small fraction of exceptionally high-resolution images (note that the considered testing sets contain images of widely varying resolution), we follow L3C in extracting 4 non-overlapping crops $x_c$ from the image $x$ such that combining $x_c$ yields $x$. We then compress the crops individually.
However, we evaluate the non-learned baselines on the full images to avoid a bias in favor of our method.

\subsection{Training Procedures}
\label{sec:trainingproc}

\paragraph{Residual Compressor}
We train for $50$ epochs on batches of 16 random $128{\times}128$ crops extracted from the training set, using the \mbox{RMSProp} optimizer~\cite{hinton2012neural}. We start with an initial learning rate (LR) of $5\textsc{e}^{-5}$, which we decay every $100\;000$ iterations by a factor of $0.75$. 
Since our Q-Classifier is trained on the output of a trained RC network, it is not available while training the RC network. Thus, we compress the training images with a random $Q$ selected from $\{12, 13, 14\}$, obtaining a pair $(x, \xl)$ for every image.

\paragraph{Q-Classifier}
Given a trained RC network, we randomly select 10\% of the training set, and compress each selected image $x$ once for each $Q \in \mathcal{Q}$, obtaining a $\xl^{(Q)}$ for each $Q\in\mathcal{Q}$. We then evaluate RC for each pair $(x, \xl^{(Q)})$ to find the optimal $Q'$ that gives the minimum bitrate for that image.
The resulting list of pairs $(x, \xl^{(Q')})$ forms the training set for the QC. 
For training, we use a standard cross-entropy loss between the softmax-normalized logits and the one-hot encoded ground truth $Q'$. We train for 11 epochs on batches of 32 random $128{\times}128$ crops, using the Adam optimizer~\cite{kingma2014adam}. We set the initial LR to the Adam-default $1\textsc{e}^{-4}$, and decay after 5 and 10 epochs by a factor of $0.25$. 

\subsection{Architecture and Training Ablations}

\paragraph{Training on Fixed $Q$}
As noted in Section~\ref{sec:trainingproc}, we select a random $Q$ during training, since QC is only available after training. We explored fixing $Q$ to one value (trying $Q \in \{12, 13, 14\}$) and found that this hurts generalization performance. This may be explained by the fact that RC sees more varied residual statistics during training if we have random $Q$'s.

\paragraph{Effect of the Crop Size}
Using crops of $128{\times}128$ to train a model evaluated on full-resolution images may seem too constraining. To explore the effect of crop size,
we trained different models, each seeing the same number of pixels in every iteration, but distributed differently in terms of batch size vs.\ crop size. We trained each model for $600\;000$ iterations, and then evaluated on the Open Images validation set (using a fixed $Q=14$ for training and testing). The results are shown in the following table and indicate that smaller crops and bigger batch-sizes are beneficial.  
    \begin{center}
    \begin{tabular}{llr}
        \toprule
        Batch Size & Crop Size & BPSP on Open Images \\
        \midrule
        16 & $128{\times}128$ & 2.854 \\
        4 & $256{\times}256$ & 2.864 \\
        1 & $512{\times}512$ & 2.877 \\
        \bottomrule
    \end{tabular}
    \end{center}

\paragraph{GDN}
We found that the GDN layers are crucial for good performance. We also explored instance normalization, and conditional instance normalization layers, in the latter case conditioning on the bitrate of BPG, in the hope that this would allow the network to distinguish different operation modes. 
However, we found that instance normalization 
is more sensitive to the resolution used for training, which led worse overall bitrates.

\section{Results and Discussion}

\subsection{Compression performance in bpsp}

We follow previous work in evaluating \emph{bits per subpixel} (Each RGB pixel has 3 subpixels), \emph{bpsp} for short, sometimes called \emph{bits per dimension}.
In Table~\ref{table:results_bpsp}, we show the performance of our approach on the described test sets. On Open Images, the domain where we train, we are outperforming all methods, including FLIF. Note that while L3C was trained on the same data set, it does not outperform FLIF. 
On the other data sets, we consistently outperform both L3C and the non-learned approaches PNG, WebP, and \jpegk.

These results indicate that our simple approach of using a powerful lossy compressor to compress the high-level image content and leverage a complementary learned probabilistic model to model the low level variations for lossless residual compression is highly effective. 
Even though we only train on Open Images, our method can generalize to various domains of natural images: mobile phone pictures (CLIC.mobile), images retouched by professional photographers (CLIC.pro), as well as high-quality images with diverse complex structures (DIV2K).

\begin{figure}[hb!]
\vspace{-1ex}
\makebox[\linewidth][c]{\includegraphics[width=1.04\linewidth]{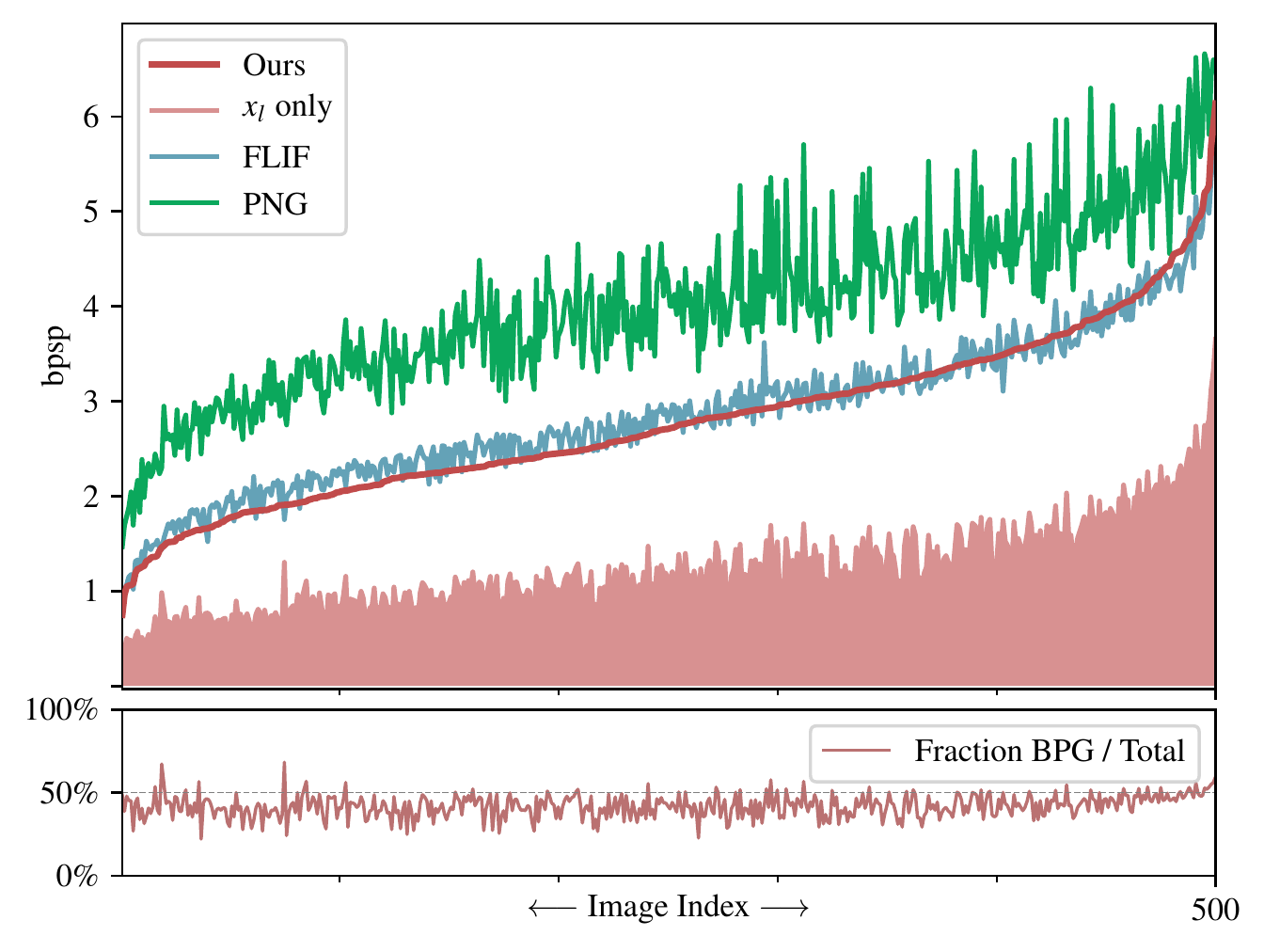}}%
\vspace{-0.5em}
\caption{\label{fig:bpp_dist}
\emph{Top}: Distribution of bpsp, on the 500 images from Open Images validation set. The images are sorted by the bpsp achieved using our approach. We show PNG and FLIF, as well as the bpsp needed to store the lossy reconstruction only (``$\xl$ only''). \emph{Bottom}: Fraction of total bits used by our approach that are used to store $\xl$. Images follow the same order as on the top panel.}
\end{figure}

In Fig.~\ref{fig:bpp_dist} we show the bpsp of each of the 500 images of Open Images, when compressed using our method, FLIF, and PNG. For our approach, we also show the bits used to store $\xl$ for each image, measured in bpsp on top (``$\xl$ only''), and as a percentage on the bottom. The percentage averages at 42\%, going up towards the high-bpsp end of the figure.
This plot shows the wide range of bpsp covered by a random set of natural images, and motivates our Q-Classifier. We can also see that while our method tends to outperform FLIF on average, FLIF is better for some high-bpsp images, where the bpsp of both FLIF and our method approach that of PNG.

\begin{figure*}[ht!]
\small
\begin{tabular}{@{}c@{\hskip 1pt}c@{\hskip 1pt}c@{\hskip 1pt}c@{\hskip 1pt}c@{\hskip 1pt}}
Input/Output $x$ & Lossy reconstruction $\xl$ & Residual $r = x - \xl$ & \multicolumn{2}{c}{Two samples from our predicted $p(r|\xl)$} \\
\includegraphics[width=0.199\textwidth]{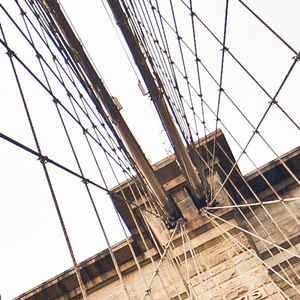} &
\includegraphics[width=0.199\textwidth]{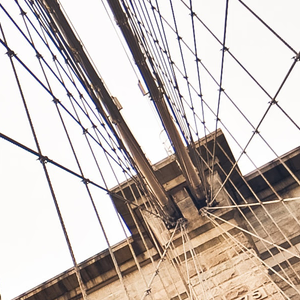} &
\includegraphics[width=0.199\textwidth]{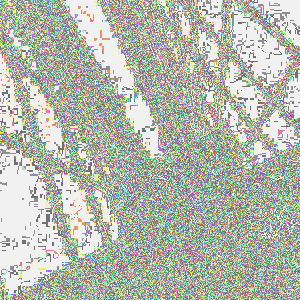} &
\includegraphics[width=0.199\textwidth]{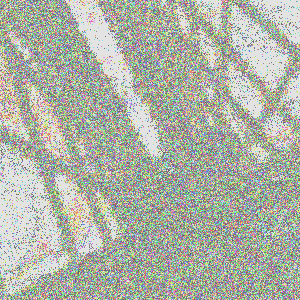} &
\includegraphics[width=0.199\textwidth]{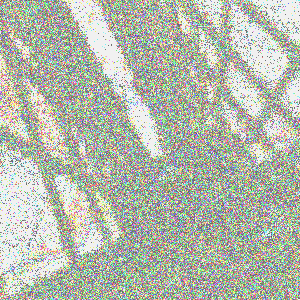} \\
\includegraphics[width=0.199\textwidth]{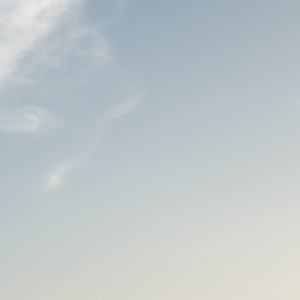} &
\includegraphics[width=0.199\textwidth]{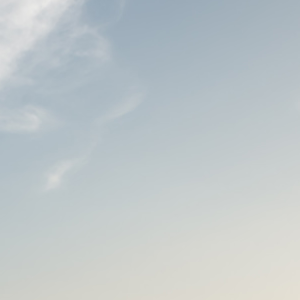} &
\includegraphics[width=0.199\textwidth]{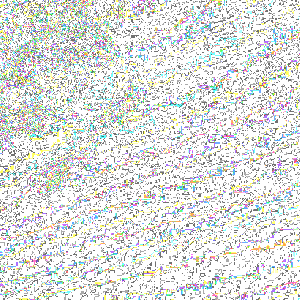} &
\includegraphics[width=0.199\textwidth]{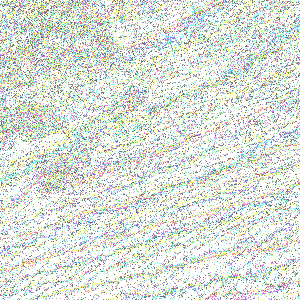} &
\includegraphics[width=0.199\textwidth]{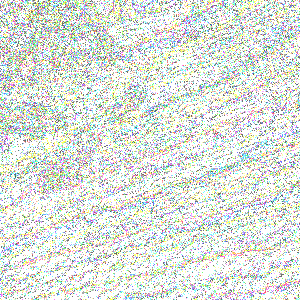} 
\end{tabular}
    \caption{\label{fig:residuals}Visualizing the learned distribution $p(r|\xl)$ by sampling from it. We compare the samples to the ground-truth target residual $r$. We also show the image $x$ that we losslessly compress as well as the lossy reconstruction $\xl$ obtained from BPG. For easier visualizations, pixels in the residual images equal to 0 are set to white, instead of gray. \emph{Best viewed on screen due to the high-frequency noise.}}
\end{figure*}

\subsection{Runtime} \label{sec:runtime}

We compare the decoding speed of RC to that of L3C for $512{\times}512$ images, using an NVidia Titan XP.
For our components: 
BPG: 163ms; RC: 166ms; arithmetic coding: 89.1ms; i.e., in a total 418ms, compared to L3C's 374ms.

QC and $\tau$-optimization are only needed for encoding. We discussed above that both components are computationally cheap. In terms of actual runtime: QC: 6.48ms; $\tau$-optimization: 35.2ms.

\subsection{Q-Classifier and $\tau$-Optimization}
In Table~\ref{table:qcomp} we show the benefits of using the Q-Classifier as well as the $\tau$-optimization. We show the resulting bpsp for the Open Images validation set (top) and for DIV2K (bottom), as well as the percentage of predicted $Q$ that are $\pm1$ away from the optimal $Q'$ (denoted ``$\pm1$ to $Q'$''), against a baseline of using a fixed $Q=14$ (the mean over QC's training set, see Section~\ref{sec:trainingproc}). 
The last column shows the required number of forward passes through RC.

\paragraph{Q-Classifier} We first note that even though the QC was only trained on Open Images (see Sec~\ref{sec:trainingproc}), we get similar behavior on Open Images and DIV2K. 
Moreover, we see that using QC is clearly beneficial over using a fixed $Q$ for all images, and 
only incurs a small increase in bpsp compared to using the optimal $Q'$
($0.18\%$ for Open Images, $0.26\%$ for DIV2K). This can be explained by the fact that QC manages to predict $Q$ within $\pm1$ of $Q'$ for 
$94.8\%$ of the images in Open Images and $90.2\%$ of the DIV2K images. 

Furthermore, the small increase in bpsp is traded for a reduction from requiring $7$ forward passes to compute $Q'$ to a single one. In that sense, using the QC is similar to the ``fast'' modes common in image compression algorithms, where speed is traded against bitrate.

\paragraph{$\tau$-Optimization} Table~\ref{table:qcomp} shows that using $\tau$-Optimization on top of QC reduces the bitrate on both testing sets. 

\paragraph{Discussion} While the gains of both components are small, their computational complexity is also very low (see Section~\ref{sec:runtime}).
As such, we found it quite impressive to get the reported gains. We believe the direction of tuning a handful of parameters post training on an instance basis is a very promising direction for image compression. One fruitful direction could be using dedicated architectures and including a tuning step end-to-end as in meta learning.

\subsection{Visualizing the learned $p(r|\xl)$}
\label{sec:visualizesamples}

\begin{table}[b]
\setlength{\tabcolsep}{5pt}
\vspace{-2ex}
    \centering
    \begin{tabular}{llllr}
        \toprule
         Data set & Setup          & bpsp  & $\pm1$ to $Q'$ & \# forward\\
        \midrule
        Open    & Optimal $Q'$    & 2.789 & 100\%  & $|\mathcal Q| = 7$  \\
        Images  & Fixed $Q=14$     & 2.801 & 82.6\% & 1 \\
                & Our QC          & 2.794 & \multirow{2}{*}{94.8\%} & 1 \\
                & Our QC + $\tau$ & 2.790 &   & 1 \\
        \midrule
        DIV2K & Optimal $Q'$    & 3.080 & 100\%  & $|\mathcal Q| = 7$  \\
              & Fixed $Q=14$     & 3.096 & 73.0\% & 1 \\
              & Our QC          & 3.088 & \multirow{2}{*}{90.2\%} & 1 \\
              & Our QC + $\tau$ & 3.079 &  & 1 \\
        \bottomrule
        \vspace{0.1ex}
    \end{tabular}
    \caption{\label{table:qcomp}
    On Open Images and DIV2K, we compare using the optimal $Q'$ for encoding images, vs.\ a fixed $Q=14$ and vs.\ using $Q$ predicted by the Q-Classifier. For each data set, the last row shows the additional gains obtained from applying the $\tau$-optimization.
    The forth column shows the percentage of predicted $Q$ that are $\pm1$ away from the optimal $Q'$ and the last column corresponds to the number of forward passes required for $Q$-optimization.
    }
\end{table}

While the bpsp results from the previous section validate the compression performance of our model, it is interesting to investigate the distribution predicted by RC. Note that we predict a mixture \emph{distribution} per pixel, which is hard to visualize directly. Instead, we sample from the predicted distribution. We expect the samples to be visually similar to the ground-truth residual $r=x-\xl$.

The sampling results are shown in Fig.~\ref{fig:residuals}, where we visualize two images from CLIC.pro with their lossy reconstructions, as obtained by BPG. We also show the ground-truth residuals $r$. Then, we show two samples obtained from the probability distribution $p(r|\xl)$ predicted by our RC network. For the top image, $r$ is in $\{-9, \dots, 9\}$, for the bottom it is in $\{-5, \dots, 4\}$ (cf.\ Fig.~\ref{fig:bpg_histo}), and we re-normalized $r$ to the RGB range $\{0, \dots, 255\}$ for visualization, but to reduce eye strain we replaced the most frequent value ($128$, i.e., gray), with white. 

We can clearly see that our approach  
i) learned to model the noise patterns
discarded by BPG
inherent with these images,
ii) learned to correctly predict a zero residual where BPG manages to perfectly reconstruct, and
iii) learned to predict structures similar to the ones in the ground-truth.

\section{Conclusion}

In this paper, we showed how to
leverage BPG to achieve
state-of-the-art results in full-resolution learned lossless image compression. Our approach outperforms L3C, PNG, WebP, and \jpegk consistently, and also outperforms the hand-crafted state-of-the-art FLIF on images from the Open Images data set. Future work should investigate input-dependent optimizations, which are also used by FLIF and which we started to explore here by optimizing the scale of the probabilistic model for the residual ($\tau$-optimization). 
Similar approaches could also be applied to latent probability models of lossy image and video compression methods.

\FloatBarrier

\newpage
\clearpage

{\small
\bibliographystyle{ieee_fullname}
\bibliography{egbib}
}

\clearpage
\onecolumn

\appendix

\setcounter{table}{0}
\renewcommand{\thetable}{A\arabic{table}}
\setcounter{figure}{0}
\renewcommand{\thefigure}{A\arabic{figure}}

\section{Learning Better Lossless Compression Using Lossy Compression -- Supplementary}\vspace{2em}

\subsection{Q-Classifier Architecture}
\label{sec:qcarch}

We show the architecture for the Q-Classifier in Table~\ref{table:qcarch}. \emph{Residual} denotes a sequence of convolution, ReLU, convolution, with a skip connection adding the input to the output (as in~\cite{he2016deep}, but without BatchNorm).

\begin{table}[h!]
    \centering
    \begin{tabular}{lrrcc}
        \toprule
        Layer &  $C_\text{in}$ & $C_\text{out}$ & Filter & Stride 
        \\ \midrule 
        Conv + ReLU  & 3 & 64 & $5{\times}5$ & 2 \\
        Conv + ReLU & 64 & 128 & $5{\times}5$ & 2 \\
        $4\times$ Residual &  128 & 128 & $3{\times}3$ \\ 
        Conv & 128 & 256 & $5{\times}5$ & 2 \\
        $4\times$ Residual & 256 & 256 & $3{\times}3$ \\ 
        Channel-Avg. & 256 & 256 & \\ 
        Linear & 256 & $|\mathcal{Q}|$ & \\ 
        \bottomrule
    \end{tabular}
    \vspace{0.2cm}
    \caption{\label{table:qcarch}Q-Classifier architecture.}
\end{table}

\subsection{BPG Performance}

\begin{figure}[h!]
\centering
\hspace{-2em}\includegraphics[width=0.5\linewidth]{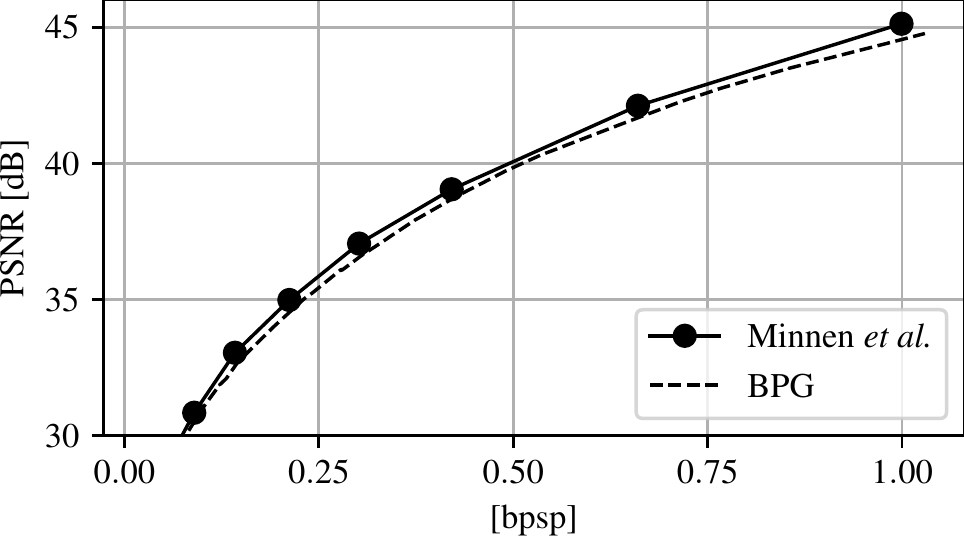}
\caption{\label{fig:bpg_perf}%
Comparing BPG to Minnen \etal~\cite{minnen2018joint}}
\end{figure}

Fig.~\ref{fig:bpg_perf} compares the performance of BPG on Kodak, in terms of PSNR, to the recent learned image compression approach from Minnen \etal~\cite{minnen2018joint}. The plot is digitized from Figure 2 in~\cite{minnen2018joint}.

\subsection{Examples from the testing sets}\label{sec:testsetexamples}

We provide additional visual examples here: 

{\small \url{https://data.vision.ee.ethz.ch/mentzerf/rc/rc_suppl_additional.pdf}}

Specifically, we show one image from each of our testing sets, alongside with the residual $r$ and a sample from $p(r)$, which is expected to be visually similar to $r$. Please refer to Section~\ref{sec:visualizesamples} for details on sampling and the visualization.
\end{document}